\documentclass{article}

\usepackage{PRIMEarxiv}

\usepackage[utf8]{inputenc} 
\usepackage[T1]{fontenc}    
\usepackage{hyperref}       
\usepackage{url}            
\usepackage{booktabs}       
\usepackage{amsfonts}       
\usepackage{nicefrac}       
\usepackage{microtype}      
\usepackage{lipsum}
\usepackage{fancyhdr}       
\usepackage{graphicx}       
\usepackage{subcaption}     
\graphicspath{{media/}}     

\title{EduSim-LLM: An Educational Platform Integrating Large Language Models and Robotic Simulation for Beginners
\thanks{\textit{\underline{Citation}}: 
\textbf{Shenqi Lu and Liangwei Zhang. EduSim-LLM: An Educational Platform Integrating Large Language Models and Robotic Simulation for Beginners. 2026}} 
}

\author{
  Shenqi Lu \\
  Hangzhou Dianzi University\\Information Engineering College \\
  Hangzhou, China \\
  \texttt{\ 229500317@hziee.edu.cn} \\
  \And
  Liangwei Zhang \\
  Hangzhou Dianzi University\\Information Engineering College \\
  Hangzhou, China \\
  \texttt{zhanglw@hziee.edu.cn} \\
}

\begin{document}
\maketitle

\begin{abstract}
In recent years, the rapid development of Large Language Models (LLMs) has significantly enhanced natural language understanding and human-computer interaction, creating new opportunities in the field of robotics. However, the integration of natural language understanding into robotic control is an important challenge in the rapid development of human-robot interaction and intelligent automation industries. This challenge hinders intuitive human control over complex robotic systems, limiting their educational and practical accessibility. To address this, we present the EduSim-LLM, an educational platform that integrates LLMs with robot simulation and constructs a language-drive control model that translates natural language instructions into executable robot behavior sequences in CoppeliaSim. We design two human-robot interaction models: direct control and autonomous control, conduct systematic simulations based on multiple language models, and evaluate multi-robot collaboration, motion planning, and manipulation capabilities. Experiential results show that LLMs can reliably convert natural language into structured robot actions; after applying prompt-engineering templates instruction-parsing accuracy improves significantly; as task complexity increases, overall accuracy rate exceeds 88.9\% in the highest complexity tests.
\end{abstract}


\section{Introduction}
In recent years, rapid advancements in human–robot interaction and artificial intelligence have enabled new opportunities for intuitive robot control. Compared with programming-based approaches, natural language control platforms provide accessible and user-friendly solutions, allowing non-experts to operate robots effectively \cite{ge2024cocobo}. Stein et al. \cite{stein2025novice} developed a browser-based robot simulation with a block-based interface to make robotics programming accessible to novice students. Souza et al. \cite{souza2018systematic} reported that Lego kits with visual block programming are often used in classroom robotics education for beginners. However, even with this progress, language-driven control still faces limitations: models may misinterpret complex instructions and may fail to operate in real time for dynamic tasks \cite{yu2026transformation}. 

Existing work on language-based robot control typically focuses on translating single-step commands for one robot. For example, Liang et al. \cite{liang2022code} used a Large Language Model (LLM) to generate Python code for a pick-and-place task, and Ahn et al. \cite{ahn2022can} employed an LLM to map commands to a fixed set of mobile-manipulation skills. These systems rely on predefined skill vocabularies or code templates and target fairly simple tasks. These systems generally lack support for multi-robot coordination or fine-grained manipulator control. In addition, interpreting or understanding the code produced by these LLMs is largely unaddressed \cite{shu2024llms}. 

Most existing educational robotics platforms do not tightly integrate LLMs with simulation environments, and their development typically involved substantial coding efforts. While high-fidelity simulators such as Gazebo provide accurate modeling, they require technical expertise for configuration \cite{gervais2021developing}. Conversely educational-oriented simulators generally are more user-friendly but suffer from practical limitation, including proprietary licenses and restricted accessibility; for instance, Robot Virtual Worlds \cite{mistretta2022virtual} and VEX Virtual \cite{robotics2022vexcode} require paid installation. As a result, current tools remain difficult to access for learners and instructors with limited programming experiments.

To address these limitations, we present EduSim-LLM, a platform that integrates LLMs with the CoppeliaSim robotics simulator for natural-language-based robot control. EduSim-LLM provides a novice-friendly interface: users type instructions in natural language and observe the corresponding simulated robots behaviors without writing any code. The system supports two interaction modes: direct operation via step-by-step commands and autonomous operation based on language. User instructions are sent to an LLM via structured prompts, and the LLM output is converted into Python control code executed in CoppeliaSim. EduSim-LLM supports multiple robot configurations, including teams of mobile robots, as well as manipulators and multiple LLM backends. We validate EduSim-LLM on concrete tasks, including multi-robot collaboration, mobile path following, and manipulator operations. Dataset-based evaluation shows that the instruction execution model maintains success rates of 100\% for simple tasks, 94.4\% for composite tasks, and 88.9\% for complex tasks. Quantitative comparisons further indicate that, for complex operations, the average human operation time under natural language control is consistently shorter than that under manual control, with time reductions exceeding 17.0 seconds across all evaluated cases.  
Our contributions are as follows: 

(1) We design a zero-coding visual platform for educational natural language robot control.

(2) We construct an educational benchmark for evaluation across instruction complexity levels.

The structure of this paper is as follows. Section 2 describes the system framework and core components of EduSim-LLM. Section 3 presents the experimental setup, instruction categories, and evaluation methodology. Section 4 concludes the paper.
 \begin{figure}[!t]
  \centering
  \includegraphics[width=0.6\textwidth]{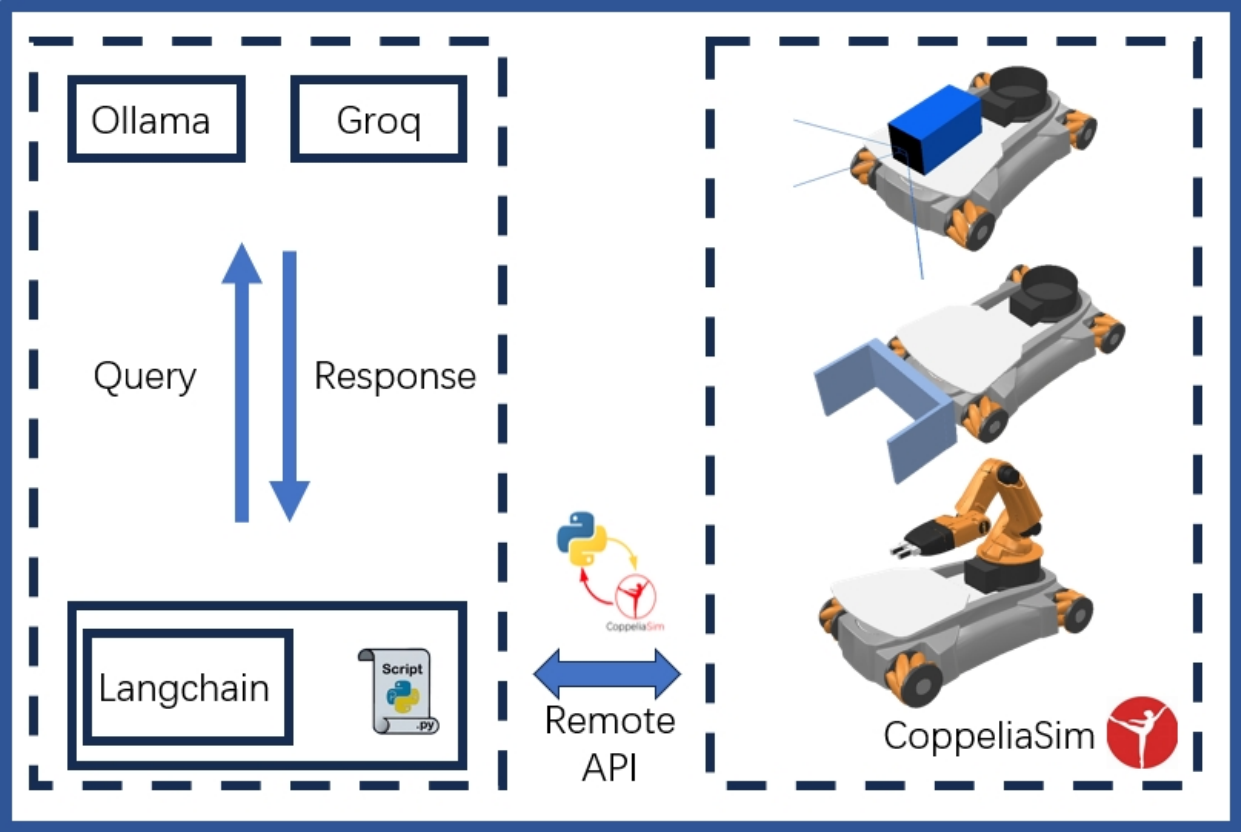}
  \caption{System architecture of EduSim-LLM integrating LLMs and CoppeliaSim}
  \label{fig: System}
\end{figure}

\section{Proposed Method}
\subsection{Design Overview}
We present EduSim-LLM, an integrated framework that bridges natural language understanding and robotic simulation through a hierarchical instruction processing pipeline the system architecture, illustrated in Fig.~\ref{fig: System}, consist of four modules: (1) Natural language interface, (2) LLM-based instruction planner, (3) Simulation control backend, and (4) User interaction frontend.

\subsection{Natural Language Interface}
\begin{figure}[!t]
  \centering
  \includegraphics[width=0.6\textwidth]{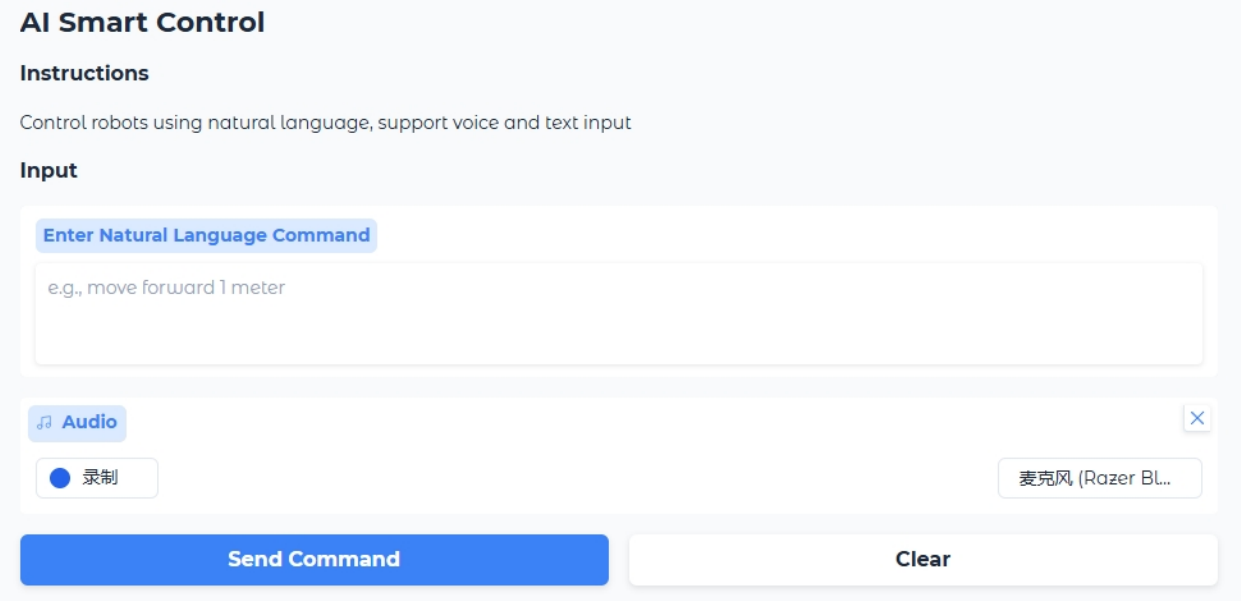}
  \caption{User interface of EduSim-LLM}
  \label{fig: Nature Language}
\end{figure}

The natural language interface functions as the input module of the EduSim-LLM system. As shown in Fig.~\ref{fig: Nature Language}, it receives user instructions in text or speech form. The input includes single-step, multi-step, and multi-agent directives. The system applies a structured prompt template to convert natural language into Python control code. This module allows users to operate robots without writing code and serves as the initial stage in the instruction-processing pipeline.

\subsection{LLM-based Instruction Planner}
The foundation of the platform is the instruction planner, which utilizes huge language models such as Groq's llama3-70b versatile and Ollama’s local llama3.1:8b. This planner translates user directives into robot-compatible action primitives and manages conditional logic among numerous robots. The planner utilizes LangChain for timely administration and streaming execution, facilitating modularity and backend interchangeability.

\subsection{Simulation Control Backend}
This component bridges high-level strategies with robotic execution via CoppeliaSim's remote API. The RobotController class provides predefined action functions for locomotion (e.g., moveForward, moveToXY, rotateToBeta) and manipulation (e.g., moveArmSequential, presetFold). The executeActionCode function dynamically parses LLM-generated commands into sequential Python calls. The backend initializes robot handles, tracks position/orientation states, implements 7-stage progressive deceleration with overshoot detection, and maintains trajectory history for optimized navigation.

\subsection{User Interaction Frontend}
The frontend is implemented using Gradio and functions as the user interaction interface. As shown in Appendix, the interface allows users to input commands, select robot instances, choose LLM backends, and execute control processes. Users can also view parsed code, execution status, and robot feedback. The frontend supports manual control modules including movement, arm joint configuration, and camera preview. These components enable command submission and system response without programming.

\section{Experiments}
\subsection{Experimental Set Up}
This experiment aims to verify the reliability and efficiency improvement of robot control code generated using a natural language model. The test platform integrates CoppeliaSim simulation software, a Python-based control interface, and three different YouBot robots. YouBot1 is equipped with a vision sensor for real-time image transmission; YouBot2 is equipped with a transport box for obstacle removal; and YouBot3 is equipped with a robotic arm for grasping and retrieving objects. All robots are equipped with differential drive bases, enabling them to perform motion and manipulation tasks. The model to be evaluated is Groq's llama-3.3-70b-versatile. The motion code generated by LLM is executed directly in the simulation environment, and performance data such as execution success rate are recorded.
\begin{figure}[!t]
    \centering
    \begin{subfigure}{0.16\textwidth}
        \centering
        \includegraphics[width=\linewidth]{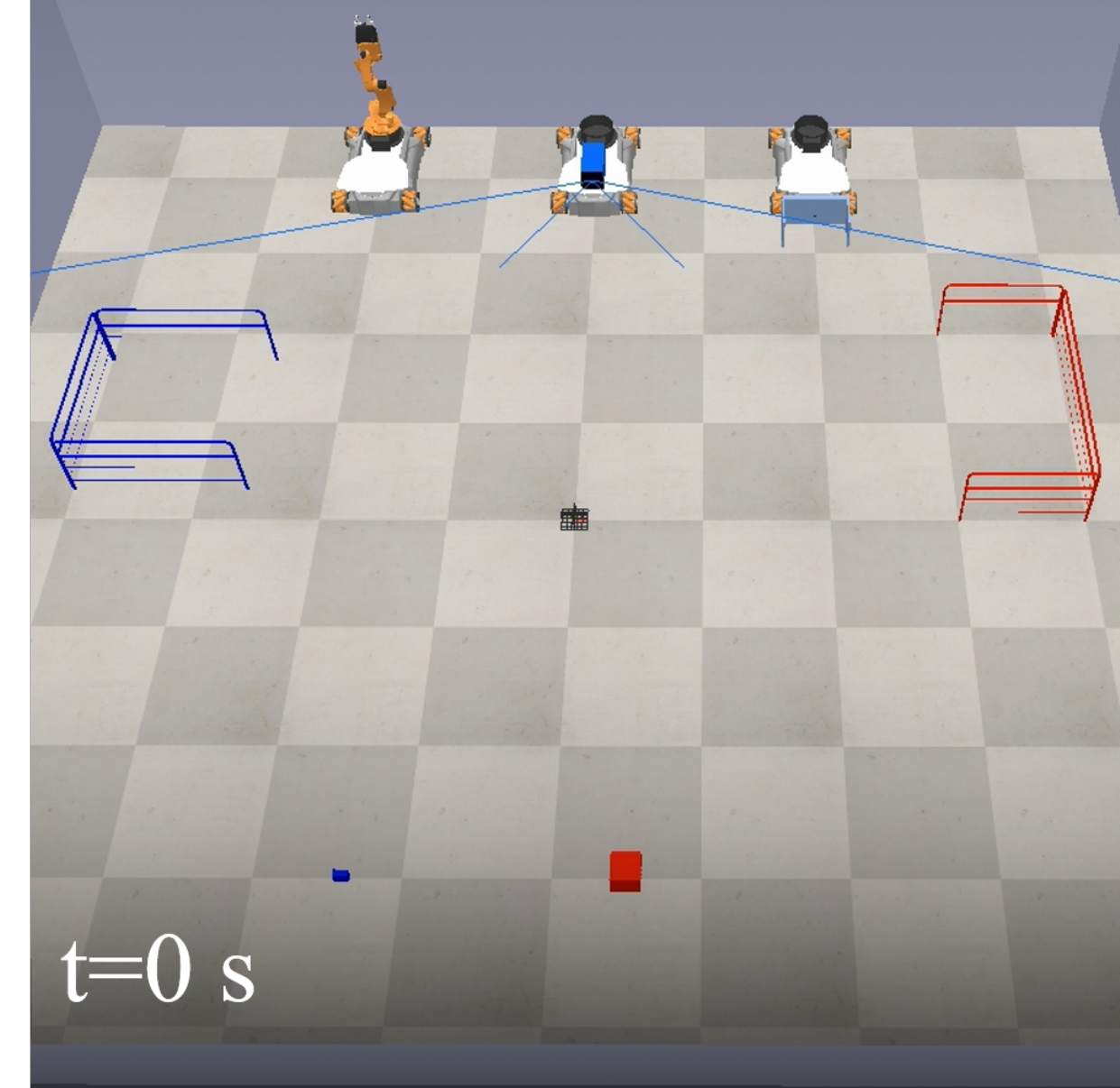}
    \end{subfigure}
    \begin{subfigure}{0.16\textwidth}
        \centering
        \includegraphics[width=\linewidth]{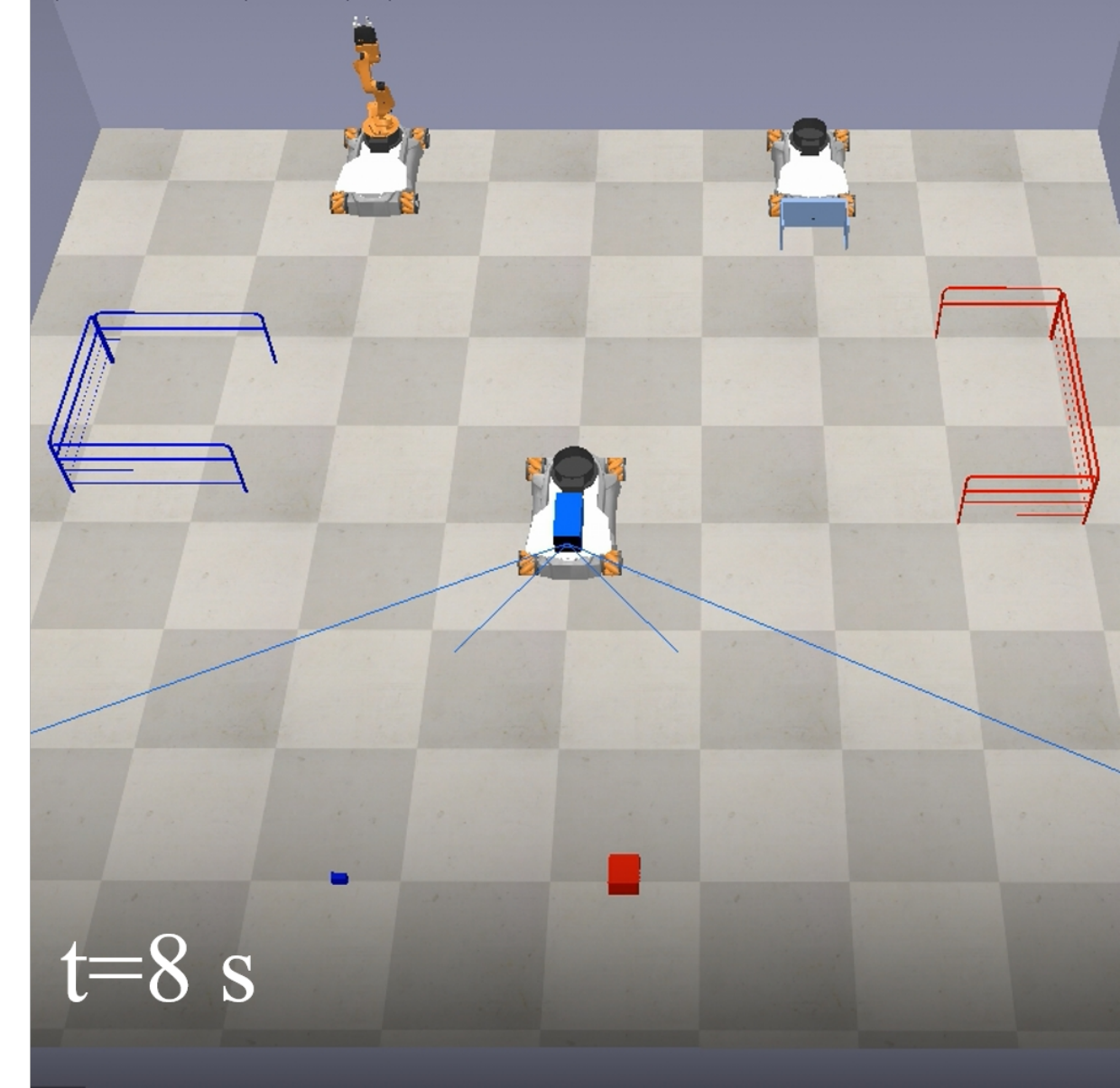}
    \end{subfigure}
    \begin{subfigure}{0.16\textwidth}
        \centering
        \includegraphics[width=\linewidth]{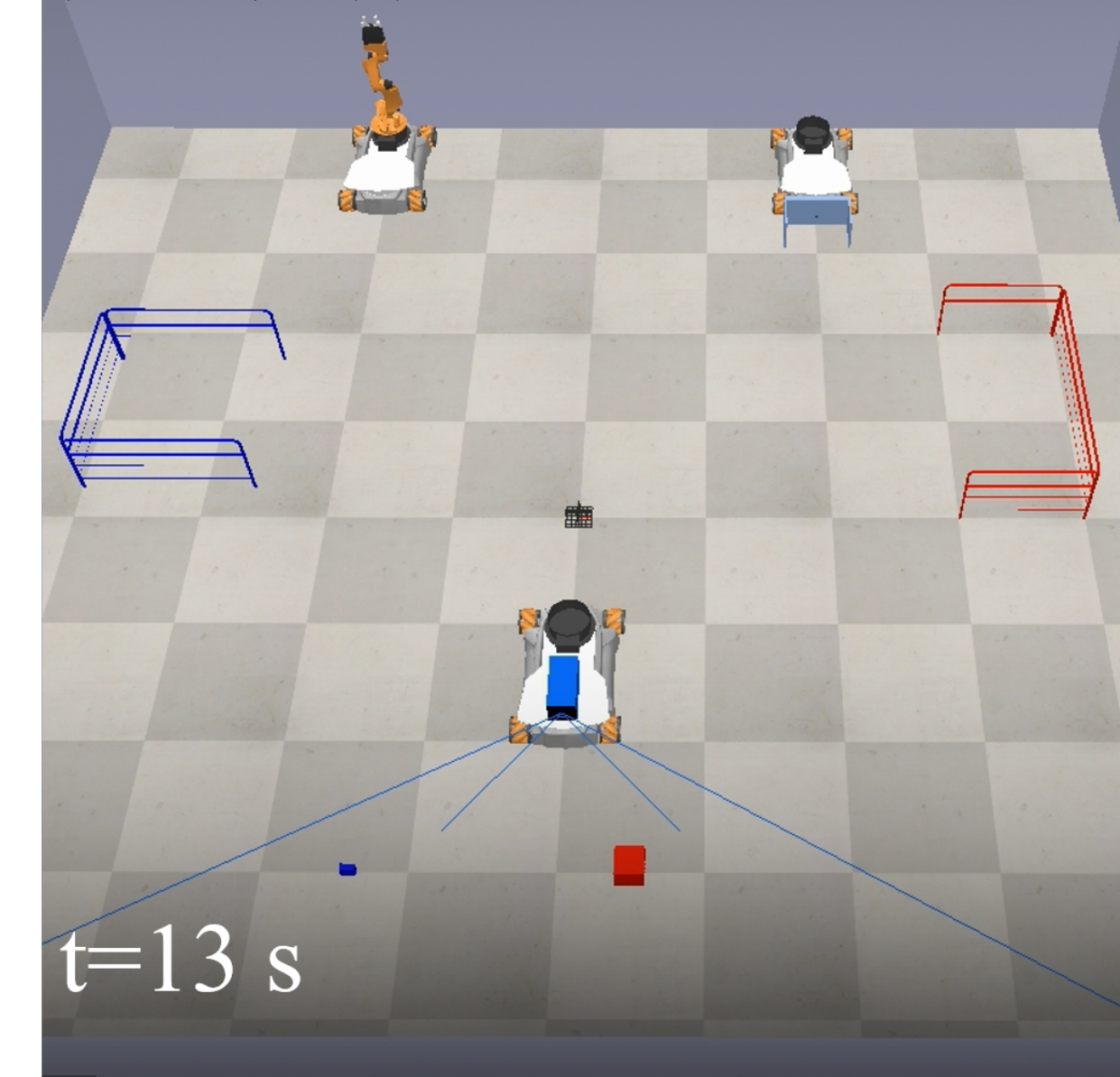}
    \end{subfigure}
    \begin{subfigure}{0.16\textwidth}
        \centering
        \includegraphics[width=\linewidth]{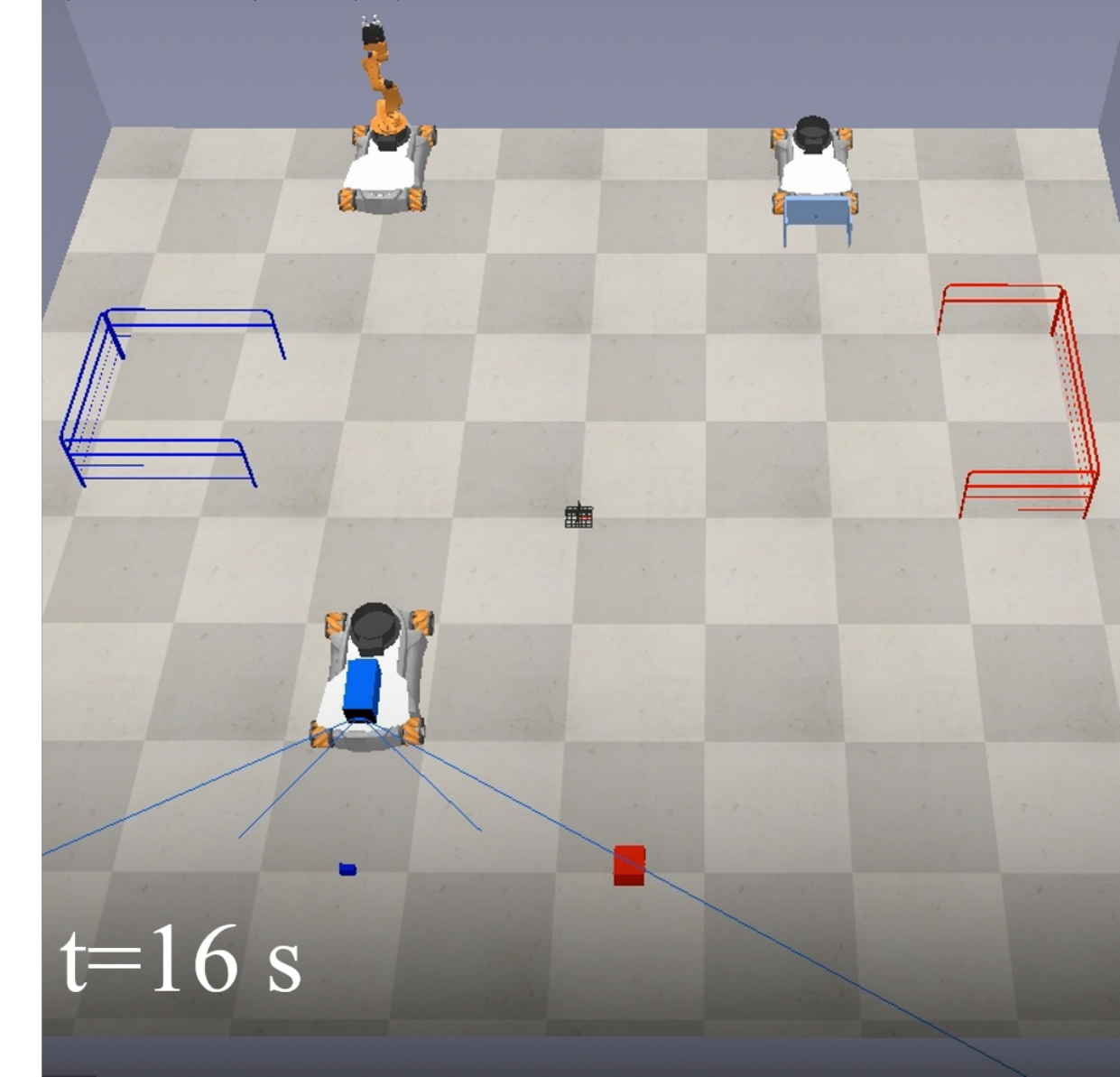}
    \end{subfigure}
    \begin{subfigure}{0.16\textwidth}
        \centering
        \includegraphics[width=\linewidth]{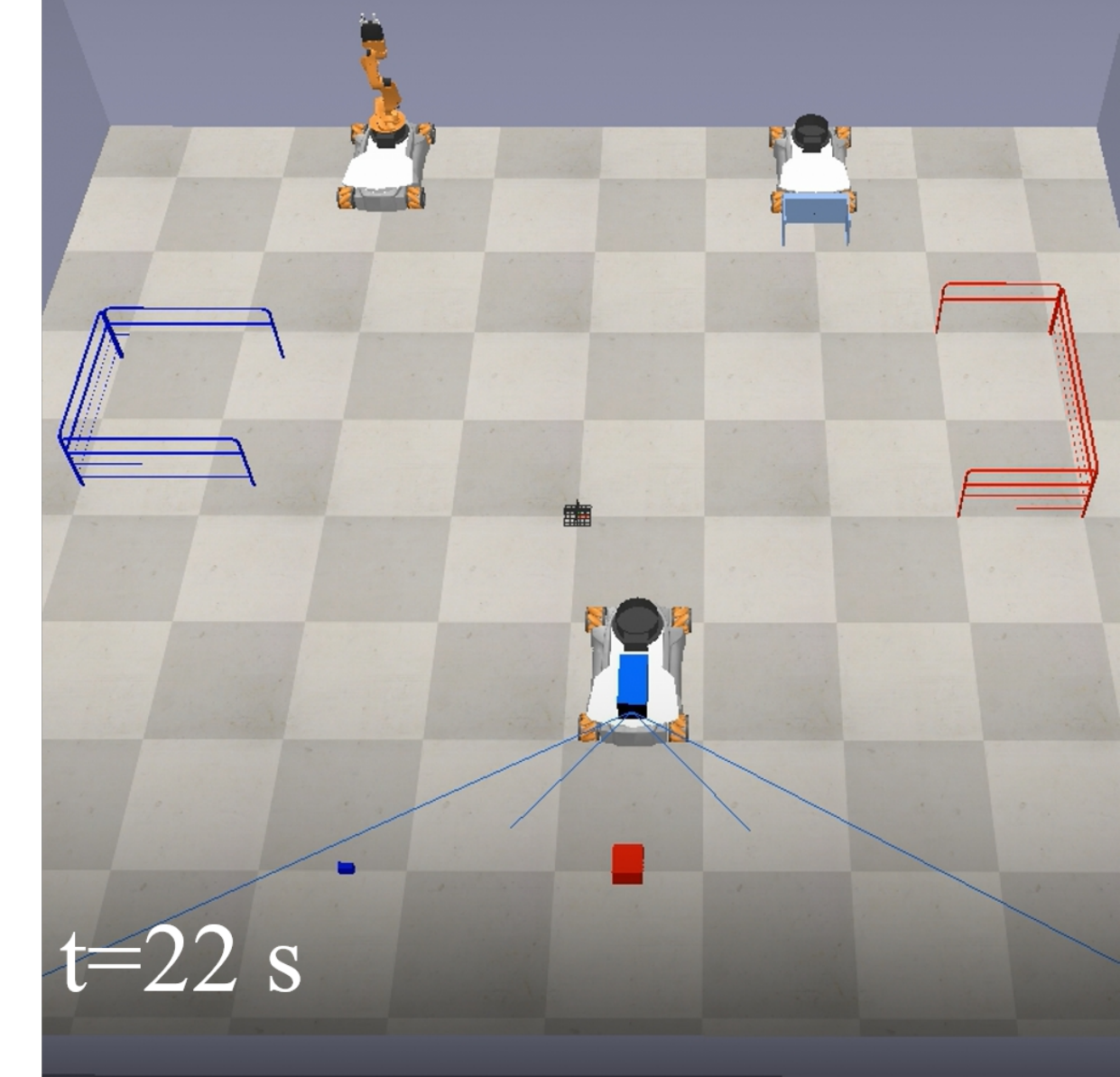}
    \end{subfigure}
    \begin{subfigure}{0.16\textwidth}
        \centering
        \includegraphics[width=\linewidth]{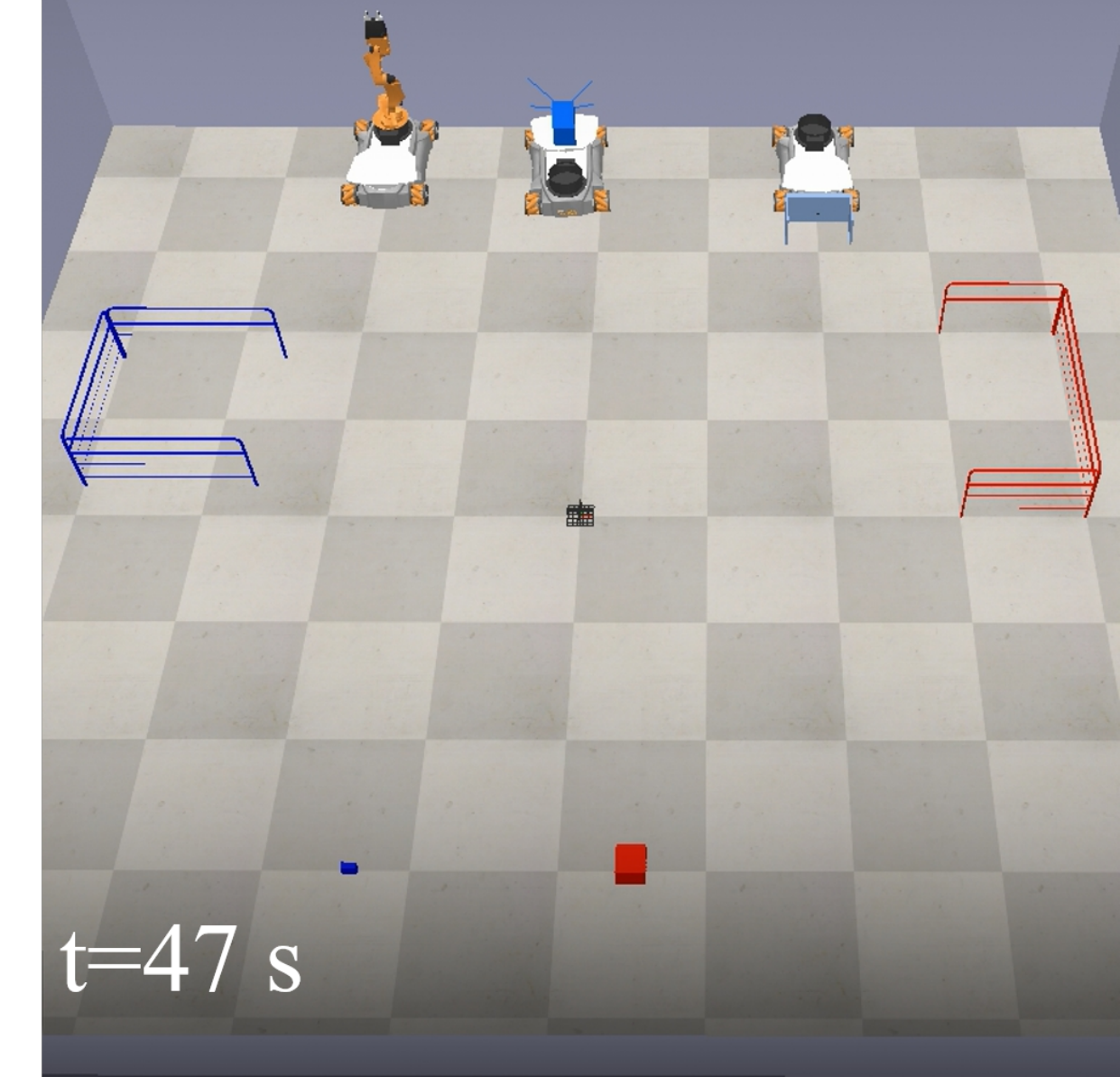}
    \end{subfigure}

    \caption{The Youbot1 robot first takes a picture at its initial position, then moves towards the negative y-axis, gets close to the target object, takes another picture, continues moving towards the target object, then moves laterally towards the negative x-axis, takes a picture before reaching the object to be grasped, then moves laterally towards the positive x-axis, takes a picture before reaching the obstacle to be removed, and finally moves back to the starting point using a combination of x and y axis movements.}
    \label{fig:Youbot1 simulation steps}
\end{figure}

\begin{figure}[!t]
    \centering
    \begin{subfigure}{0.16\textwidth}
        \centering
        \includegraphics[width=\linewidth]{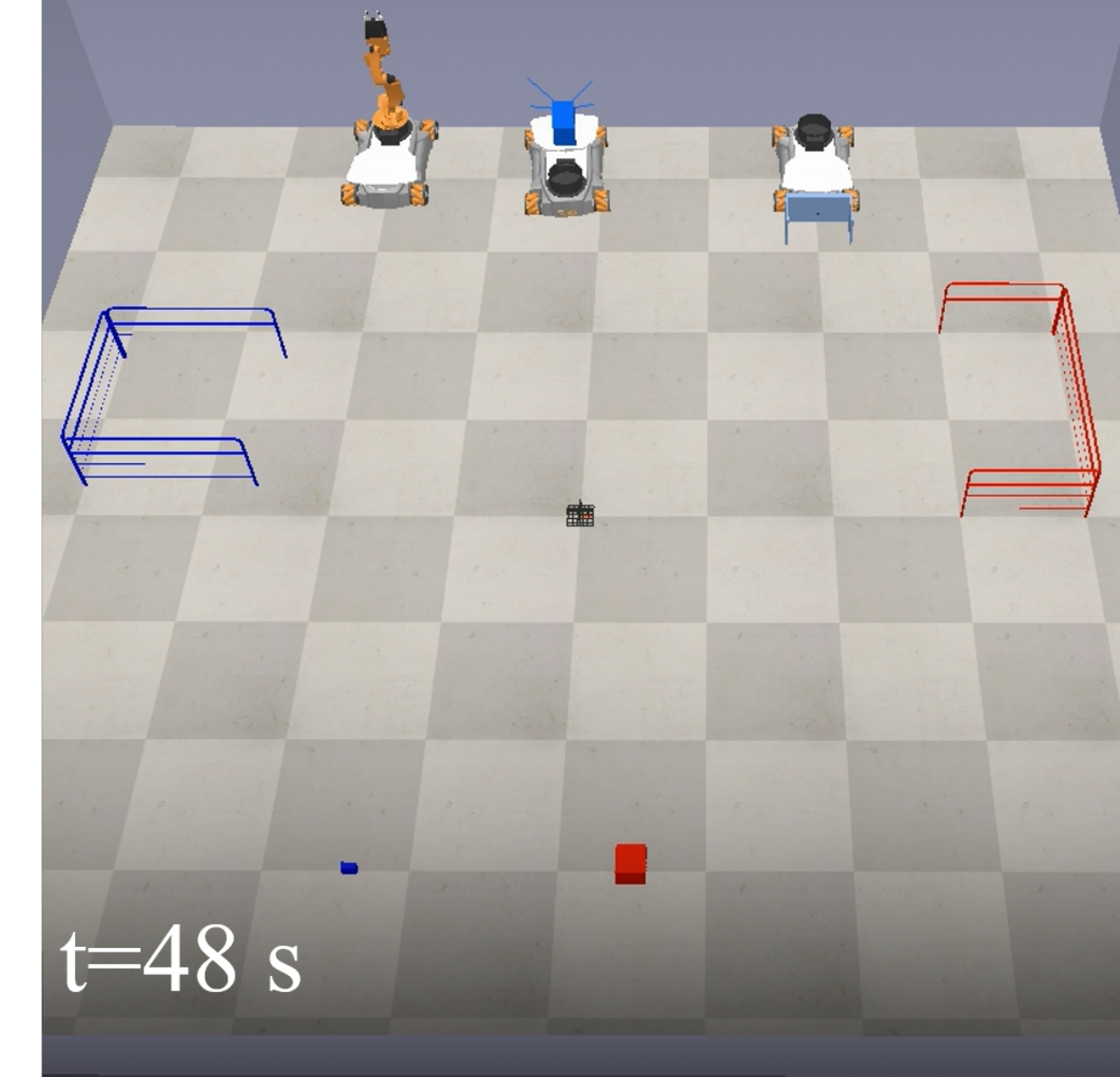}
    \end{subfigure}
    \begin{subfigure}{0.16\textwidth}
        \centering
        \includegraphics[width=\linewidth]{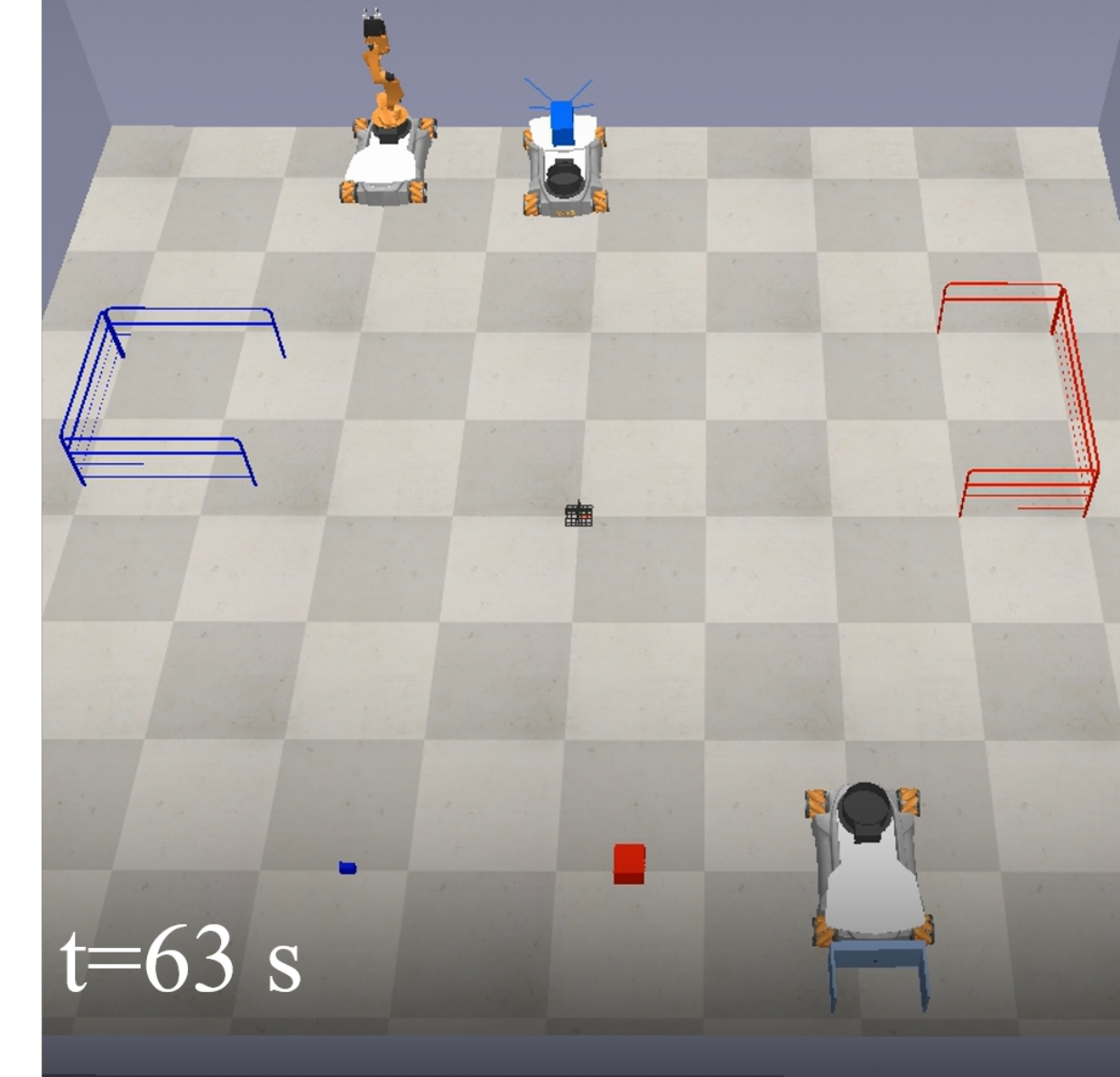}
    \end{subfigure}
    \begin{subfigure}{0.16\textwidth}
        \centering
        \includegraphics[width=\linewidth]{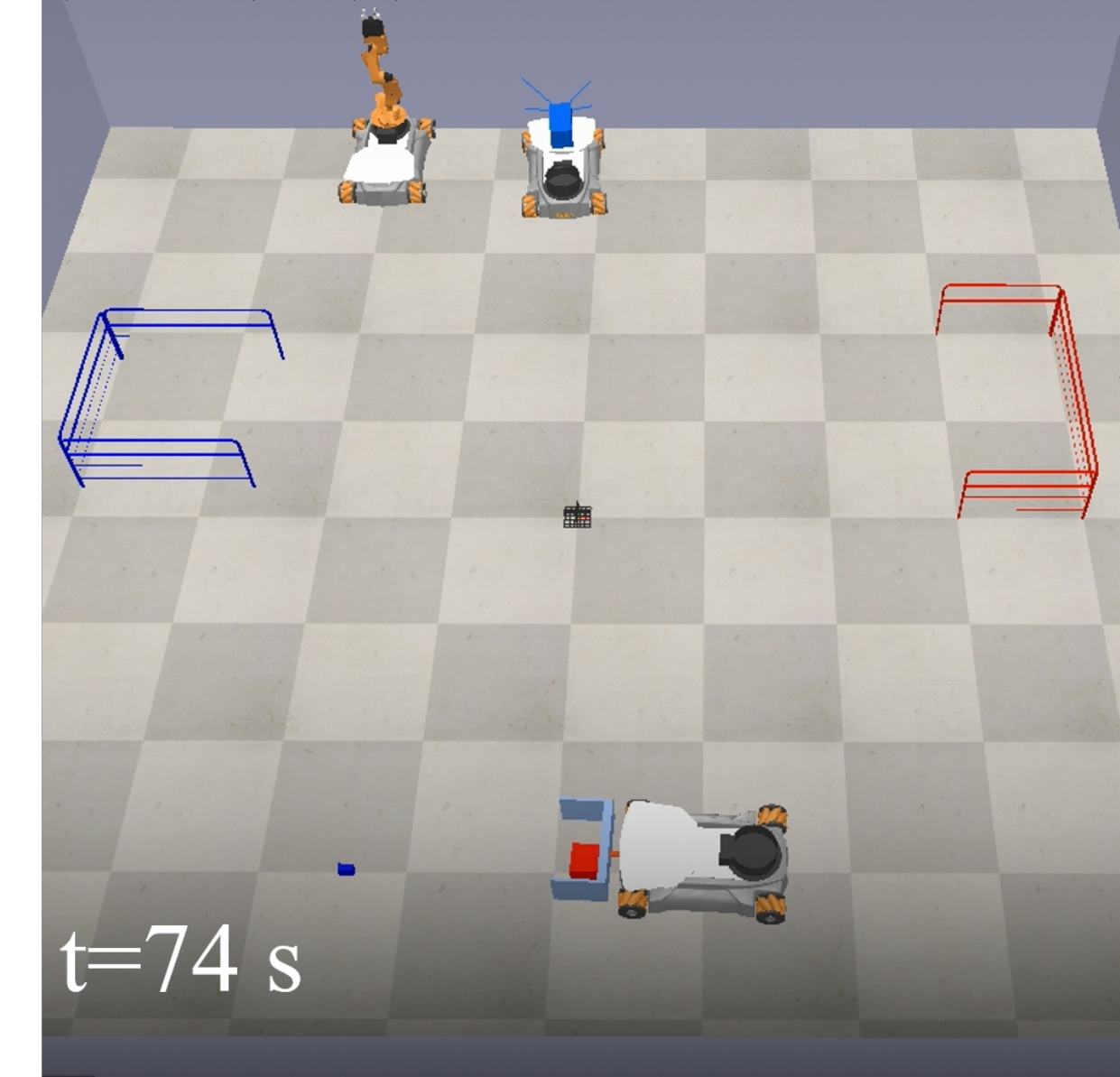}
    \end{subfigure}
    \begin{subfigure}{0.16\textwidth}
        \centering
        \includegraphics[width=\linewidth]{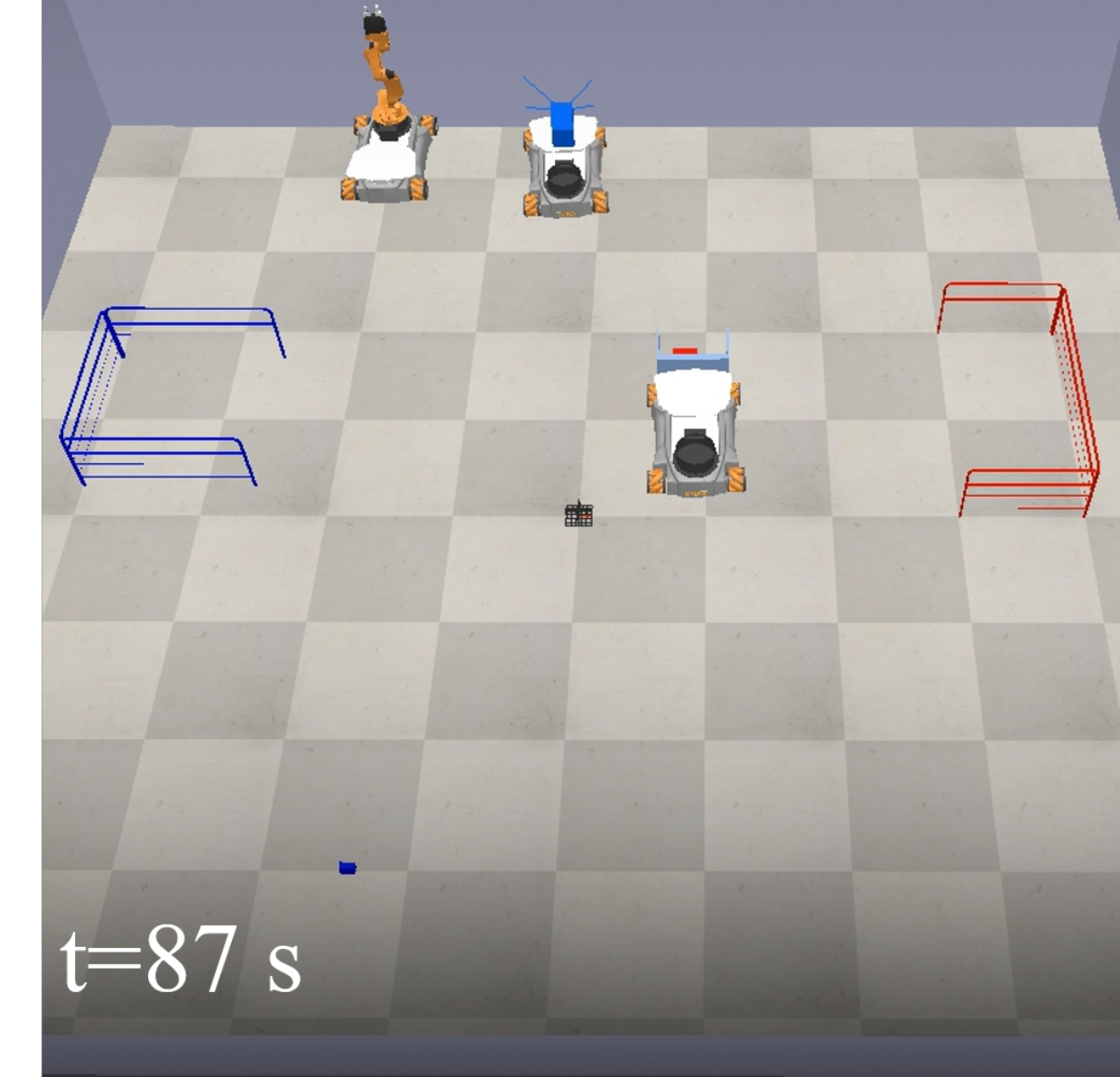}
    \end{subfigure}
    \begin{subfigure}{0.16\textwidth}
        \centering
        \includegraphics[width=\linewidth]{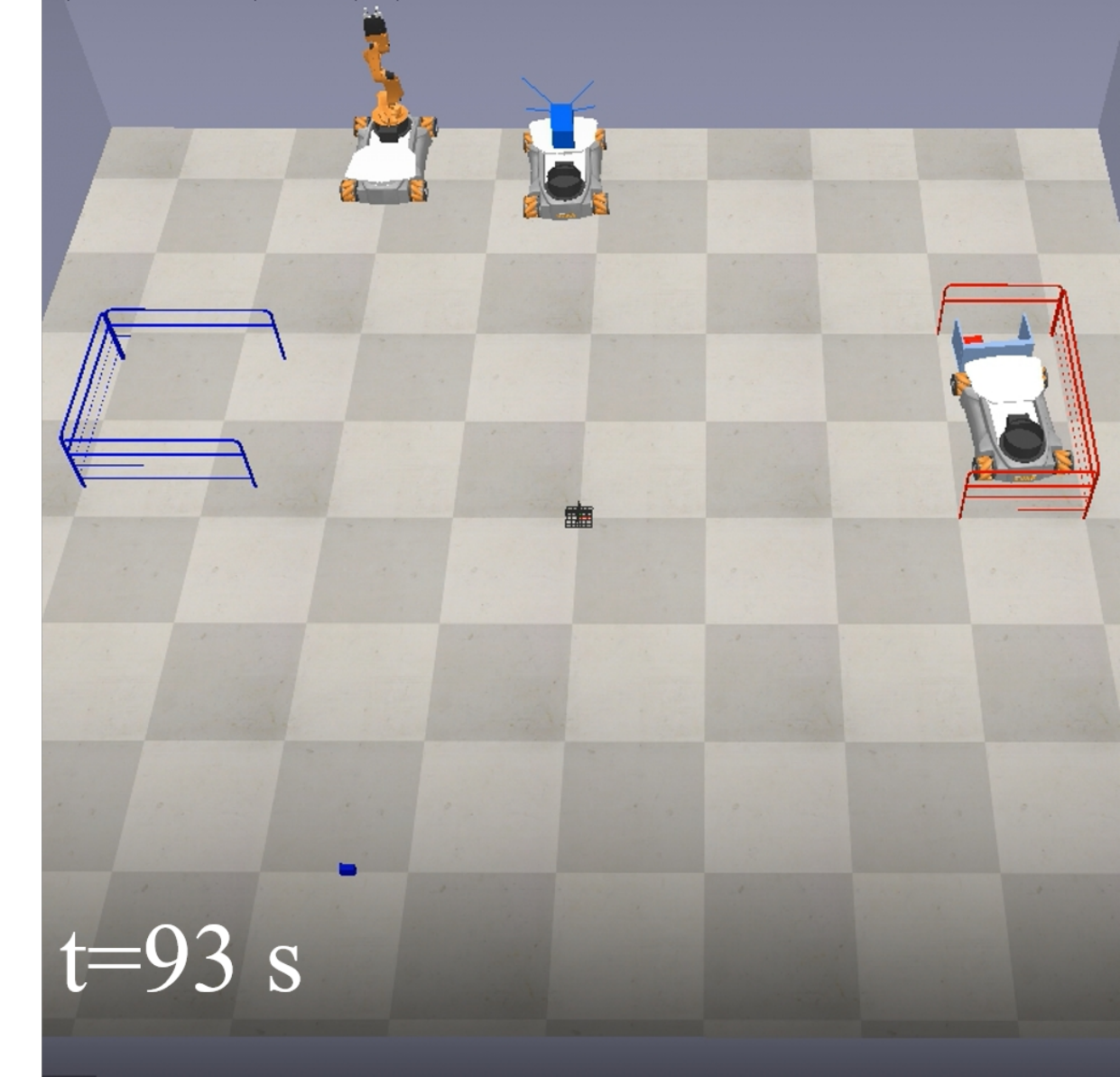}
    \end{subfigure}

    \caption{The Youbot2 robot starts from its initial position and moves in the negative y-axis direction to the obstacle to be removed. It then performs a combined x and y-axis movement to collect the obstacle into the fixed frame at the front of the robot and moves it to the designated parking area. Finally, it moves laterally in the positive x-axis direction into the parking area to complete the removal task.}
    \label{fig:Youbot2 simulation steps}
\end{figure}

\begin{figure}[!t]
    \centering
    \begin{subfigure}{0.16\textwidth}
        \centering
        \includegraphics[width=\linewidth]{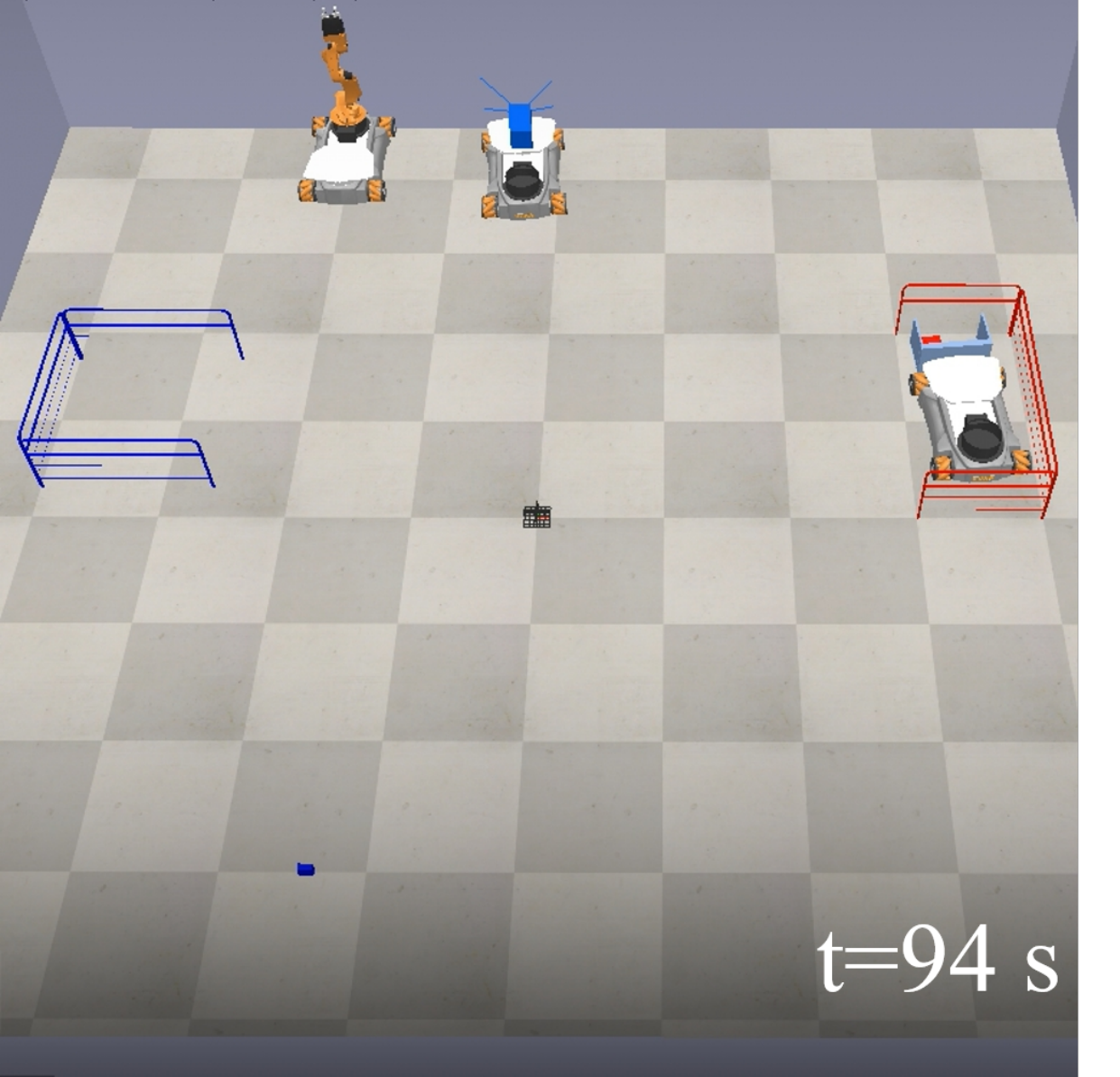}
    \end{subfigure}
    \begin{subfigure}{0.16\textwidth}
        \centering
        \includegraphics[width=\linewidth]{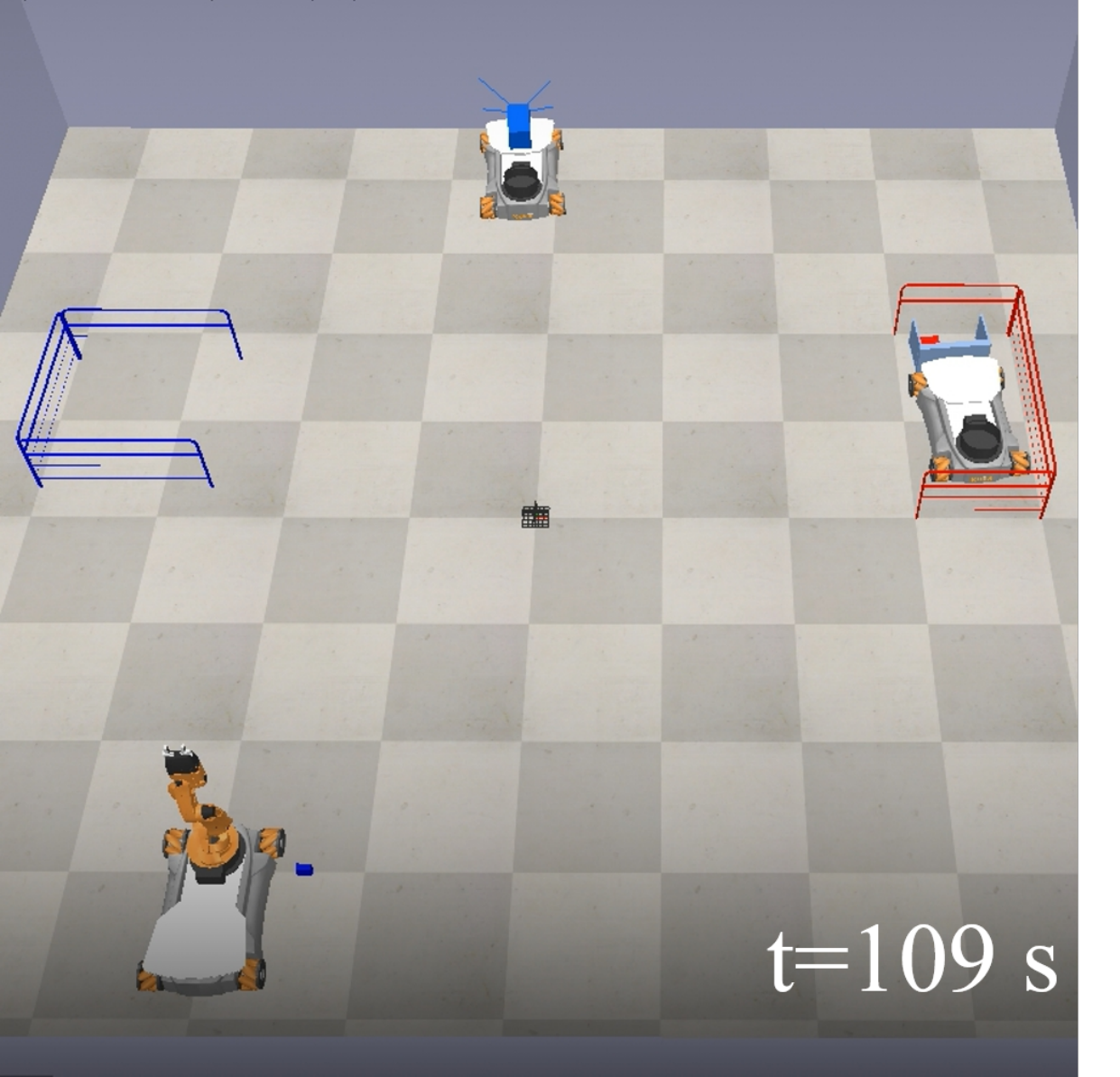}
    \end{subfigure}
    \begin{subfigure}{0.16\textwidth}
        \centering
        \includegraphics[width=\linewidth]{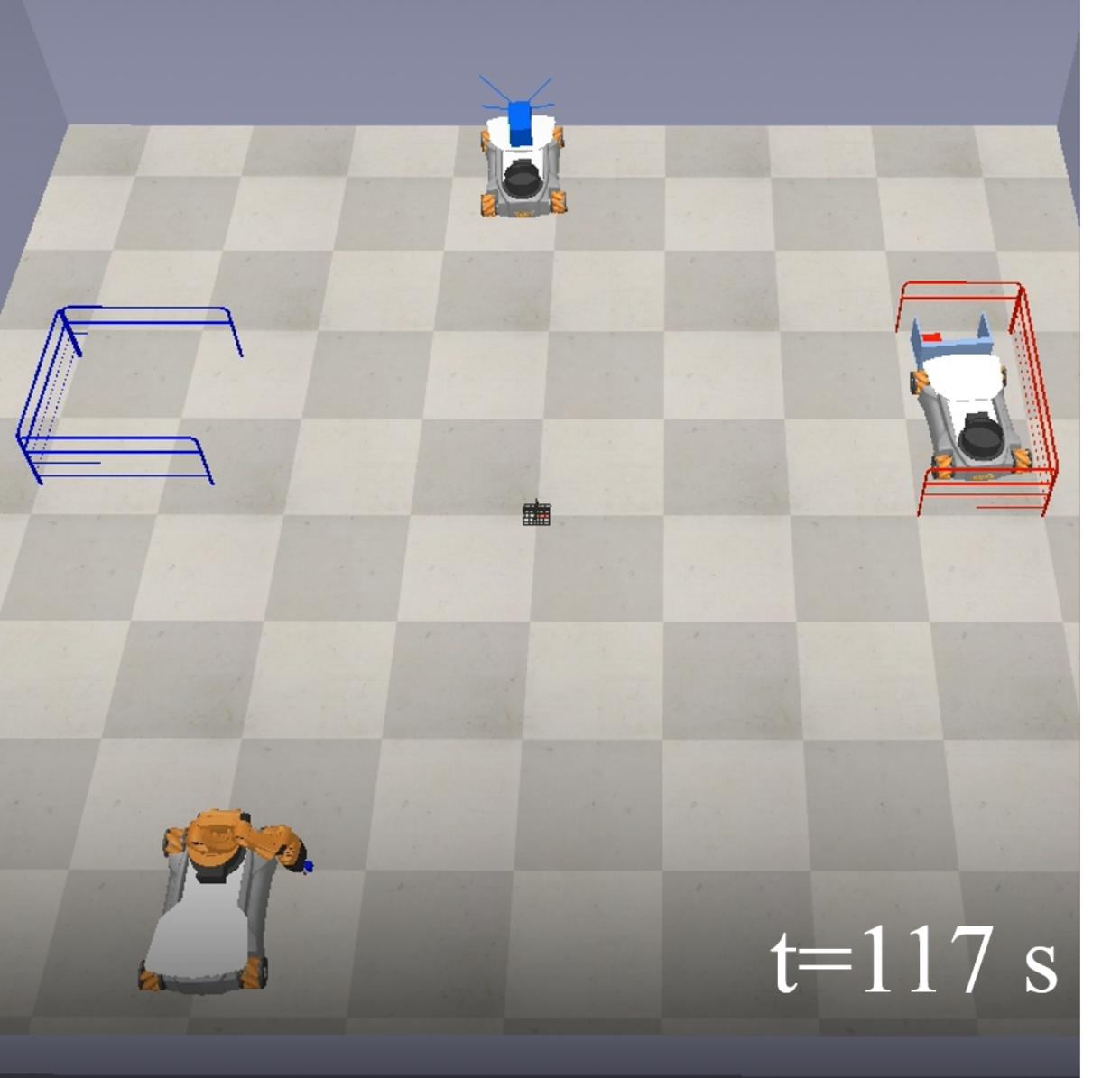}
    \end{subfigure}
    \begin{subfigure}{0.16\textwidth}
        \centering
        \includegraphics[width=\linewidth]{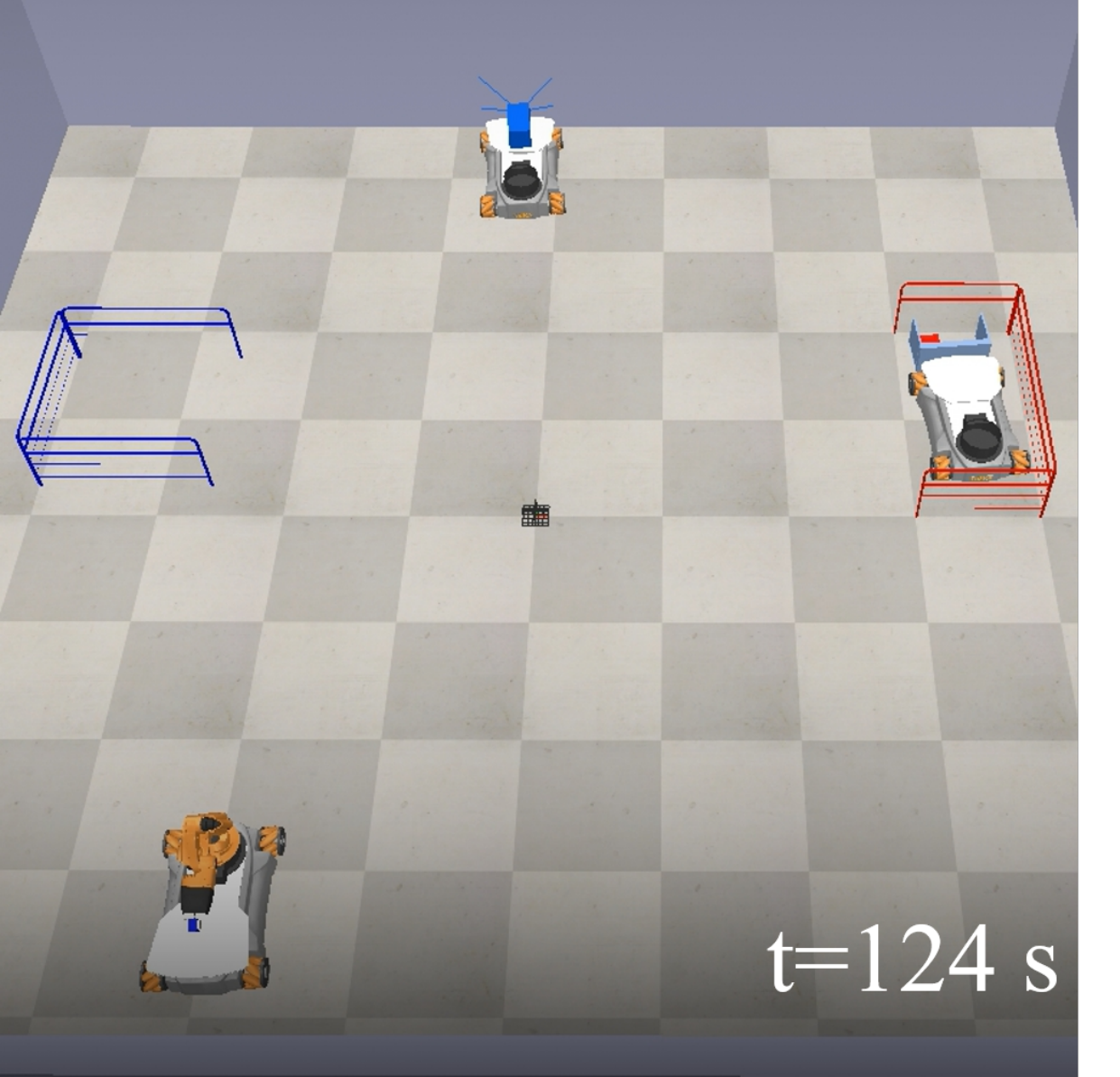}
    \end{subfigure}
    \begin{subfigure}{0.16\textwidth}
        \centering
        \includegraphics[width=\linewidth]{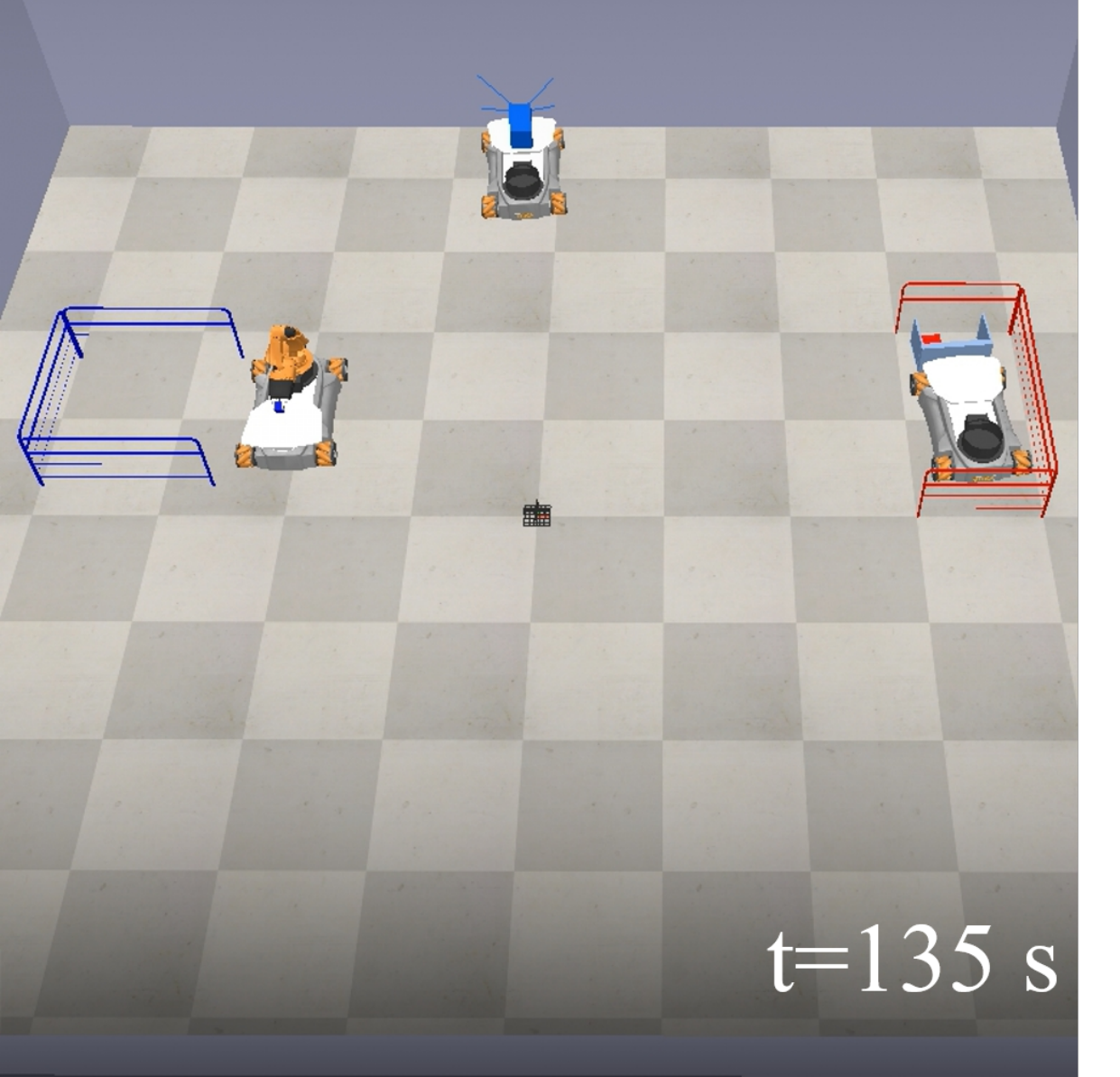}
    \end{subfigure}
    \begin{subfigure}{0.16\textwidth}
        \centering
        \includegraphics[width=\linewidth]{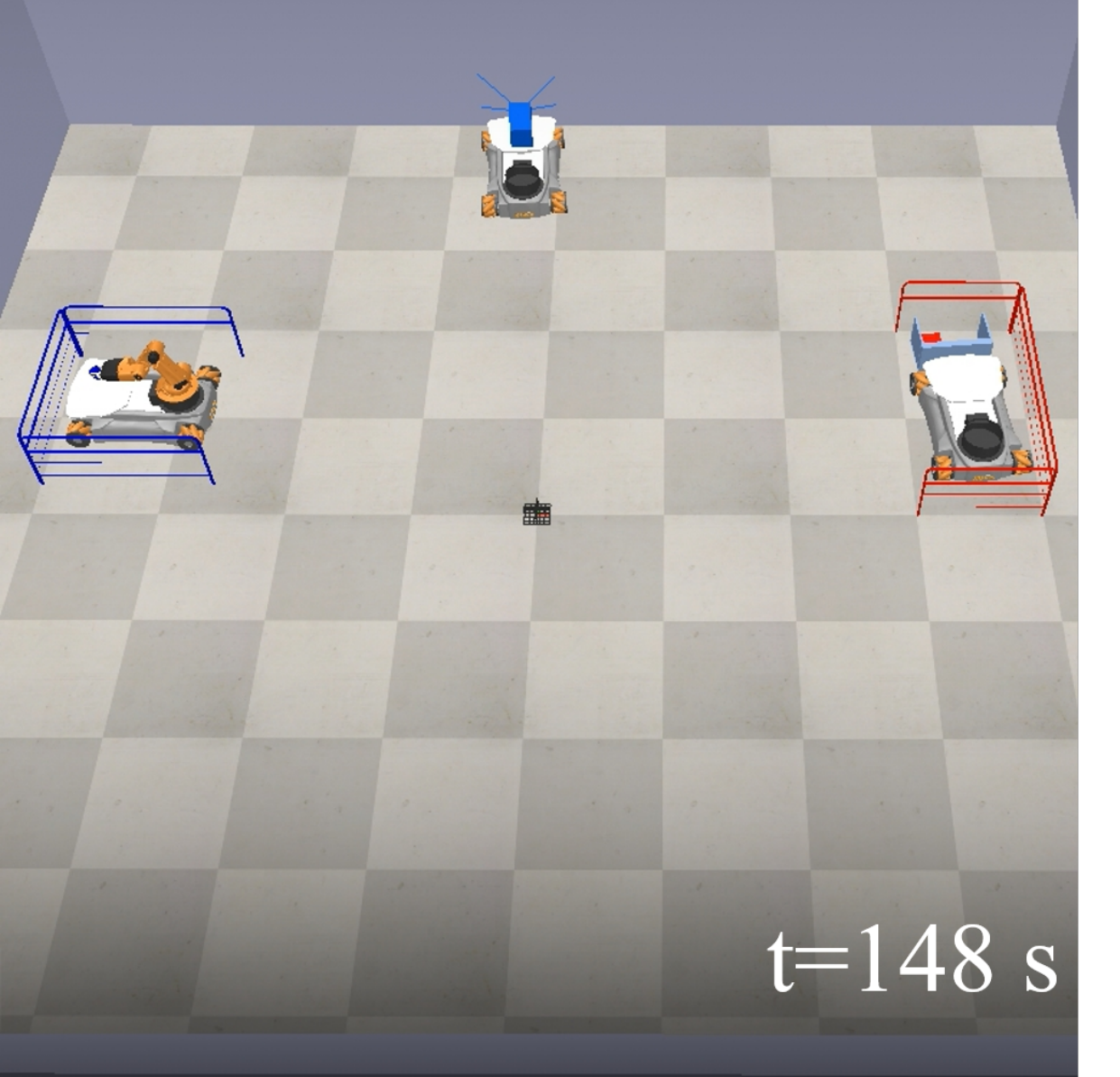}
    \end{subfigure}

    \caption{The Youbot3 robot starts from its initial position and moves in the negative y-axis direction to the object to be gripped. It then uses its robotic arm to grip and retrieve the object. After retrieving the object, it moves in the positive y-axis direction to the designated parking area, and finally moves in the negative x-axis direction into the parking area to complete the gripping task.}
    \label{fig:Youbot3 simulation steps}
\end{figure}

\subsection{Experiment Based On Dataset}
The experiment evaluates LLM-generated robot control code under different instruction difficulty levels. The test set contains 108 instances, including 36 simple, 36 composite, and 36 complex instructions. Each instance consists of a natural language command, the Python control code generated by the LLM, and a difficulty label. The definition of each difficulty level and the number of instances are shown in Table~\ref{tab:grouping_and_evaluation}.

For each test instruction, the executeActionCode function executes the generated script in the simulator and logs whether the execution succeeds and the runtime of each trial. Aggregated over 36 cases per group, the success rates for simple, composite, and complex instructions are 100\%, 94.4\%, and 88.9\%, respectively, as shown in Fig.~\ref{Fig:Experiment} (B). The success rate decreases as the instruction difficulty level increases from simple to composite and complex.

Building on the above dataset-based evaluation, a multi-robot experiment was conducted with the natural language input “all robots start action”. The system sends this text to the large language model through the prompt template, and the model decomposes it into sequential action codes for YouBot1, YouBot2, and YouBot3. YouBot1 executes a nine-step action sequence composed of alternating calls to capturePhoto, moveToY, and moveLateral, and finally calls moveToXY with the xFirst option to return to the starting point, thereby completing four image acquisition and corresponding position-adjustment subtasks. YouBot2 executes a three-step action sequence consisting of moveToY motion along the negative y-axis, moveToXY with the xFirst option to transport the obstacle, and a final lateral motion via moveLateral to reach the parking area, thereby completing the position-adjustment task along the transport path. YouBot3 executes a six-step action sequence in which moveToY is used to approach the object, presetFold and presetExtend configure the arm posture, closeGripper performs the grasping operation, moveToY transports the object along the positive y-axis, and moveToXWithRotation moves along the x-axis after rotation to reach the parking area, thereby completing the task of grasping the object and transporting it to the target location. The execution processes of YouBot1, YouBot2, and YouBot3 under this command are illustrated in Fig.~\ref{fig:Youbot1 simulation steps}, Fig.~\ref{fig:Youbot2 simulation steps}, and Fig.~\ref{fig:Youbot3 simulation steps}, respectively.
\begin{table*}[!t]
\caption{Experiment Grouping and Evaluation Variables}
\label{tab:grouping_and_evaluation}
\centering
\begin{small}
\begin{tabular}{@{}l p{0.3\textwidth} c l@{}}
\toprule
\textbf{Difficulty} & \textbf{Description}                                   & \textbf{Total Number} & \textbf{Evaluation Indicators}               \\
\midrule
Simple    & Basic single operations or 2-step sequences involving movement, arm presets, or gripper control                                & 36                     & Success rate, failure reasons                 \\
Composite & Combined 3-4 step sequences integrating navigation with arm manipulation or multiple directional movements                                & 36                     & Success rate, failure reasons                 \\
Complex   & Advanced 5-6 step sequences coordinating multiple systems including precise arm positioning, gripper control, and navigation                                & 36                     & Success rate, failure reasons\\
\bottomrule
\end{tabular}
\end{small}
\end{table*}

\subsection{Comparative Experiment Of Manual Control And Natural Language Control}
This subsection compares manual control and natural language control for complex tasks. Five tasks are selected from the complex group, and each task is executed once under two conditions: manual control and natural language control based on code generated by a large language model. For each trial, the timing starts when the user begins entering the command and ends when the command is sent to the simulator. The recorded time corresponds to the human operation time before command execution.

\begin{figure}[!t]
  \centering
  \includegraphics[width=1\textwidth]{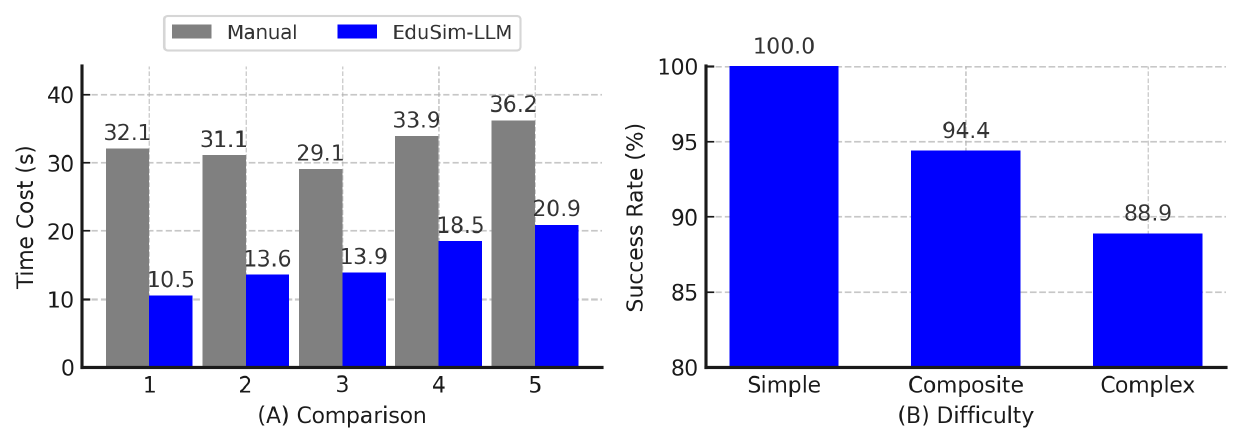}
  \caption{(A) Comparison of time costs between manual control and natural language control. Natural language control exhibits lower average execution latency than manual control across tasks, indicating improved interaction efficiency. (B) Execution success rate across different instruction difficulty levels. Each difficulty category includes 36 test cases. Although the success rate decreases as task complexity increases, the lowest rate remains at 88.9\%, demonstrating robust performance across varying instruction types.}
  \label{Fig:Experiment}
\end{figure}

Under the manual control condition, the operator issues a sequence of low-level control commands via keyboard and mouse input to specify the motion of the mobile base, the manipulator, and camera capture. Under the natural language control condition, the user inputs a natural language instruction, the system sends this instruction to the large language model through the prompt template, and the Python control code generated by the model is sent as a single unit to the simulator for execution. For these five complex tasks, the human operation time under the manual control condition exceeds 29 s for all tasks, while the human operation time under the natural language control condition is less than 21 s for all tasks, as illustrated in Fig.~\ref{Fig:Experiment} (A). For all five tasks, natural language control requires less human operation time than manual control. This comparative result indicates that, given a pre-constructed control function library and prompt template, natural language control can reduce the time users spend preparing and sending commands for complex robot operations.

\section{Conclusions}
This study proposed EduSim-LLM, an educational platform that integrates large language models with robotic simulation to enable natural-language-based robot control. The system architecture combines a structured instruction interface, a large language model planner, a simulation control backend, and a user-facing frontend. Through experimental evaluations on three types of robots and 108 instruction cases with varying complexity, the results demonstrate that LLMs can effectively translate natural language commands into executable control code. The system achieved a 100\% success rate for simple tasks, 94.4\% for composite tasks, and 88.9\% for complex tasks. Multi-robot cooperative experiments further validated the model's capability to generate sequential actions across agents. Comparative experiments with manual control showed that natural language control significantly reduces user operation time for complex tasks. The integration of structured prompts and a pre-constructed function library improves parsing reliability and system responsiveness. These findings confirm the feasibility and efficiency of using LLMs for educational robotic control in simulation environments.

\bibliographystyle{unsrt}  
\bibliography{references}  

\newpage
\appendix
\section{Pictures' Appendix For User Interaction Frontend}
\begin{figure}[ht]
  \centering
  \includegraphics[width=1\textwidth]{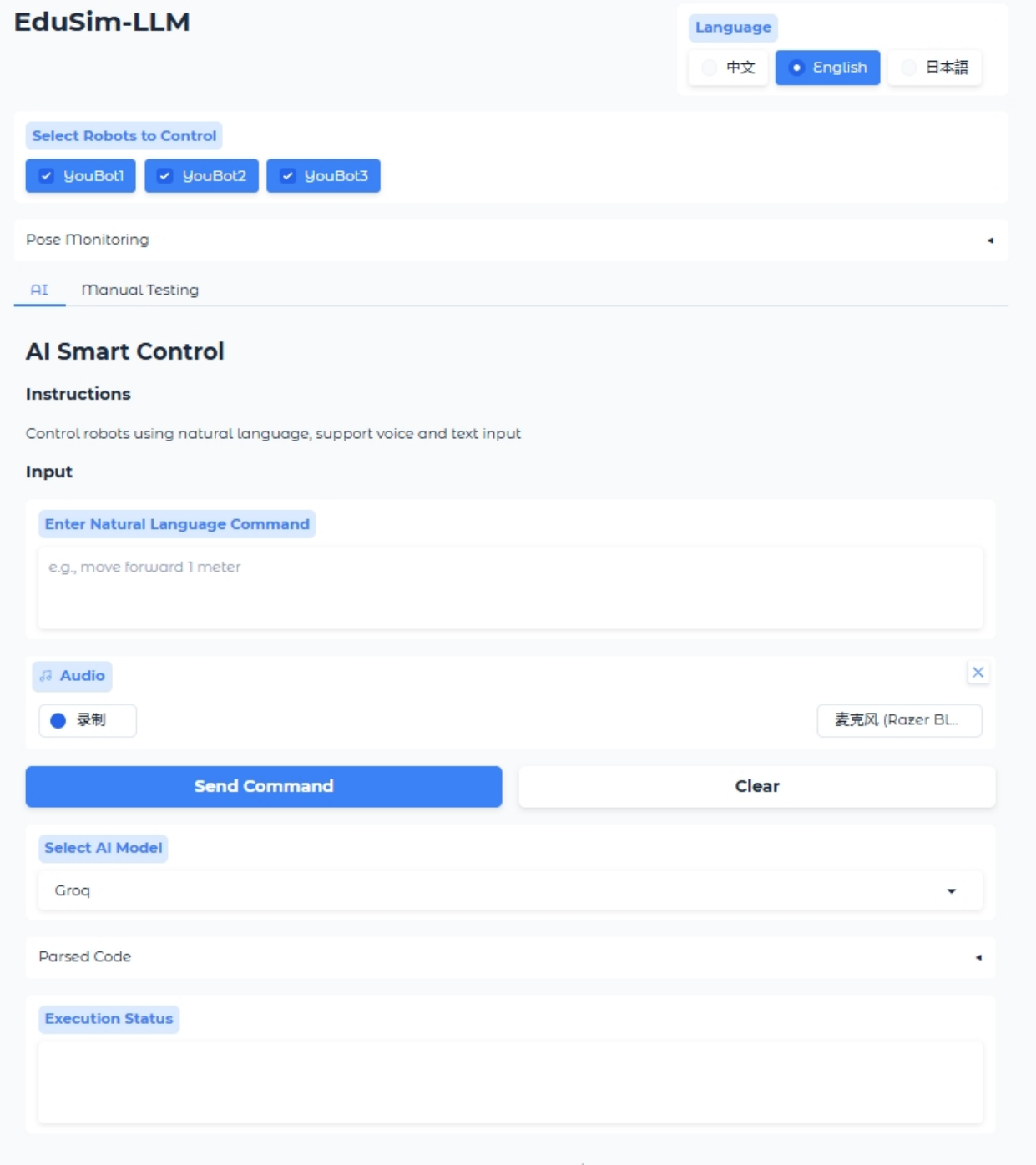}
  \caption{Natural language intelligent control dashboard}
  \label{fig: Frontend1}
\end{figure}

\begin{figure}[ht]
  \centering
  \includegraphics[width=1\textwidth]{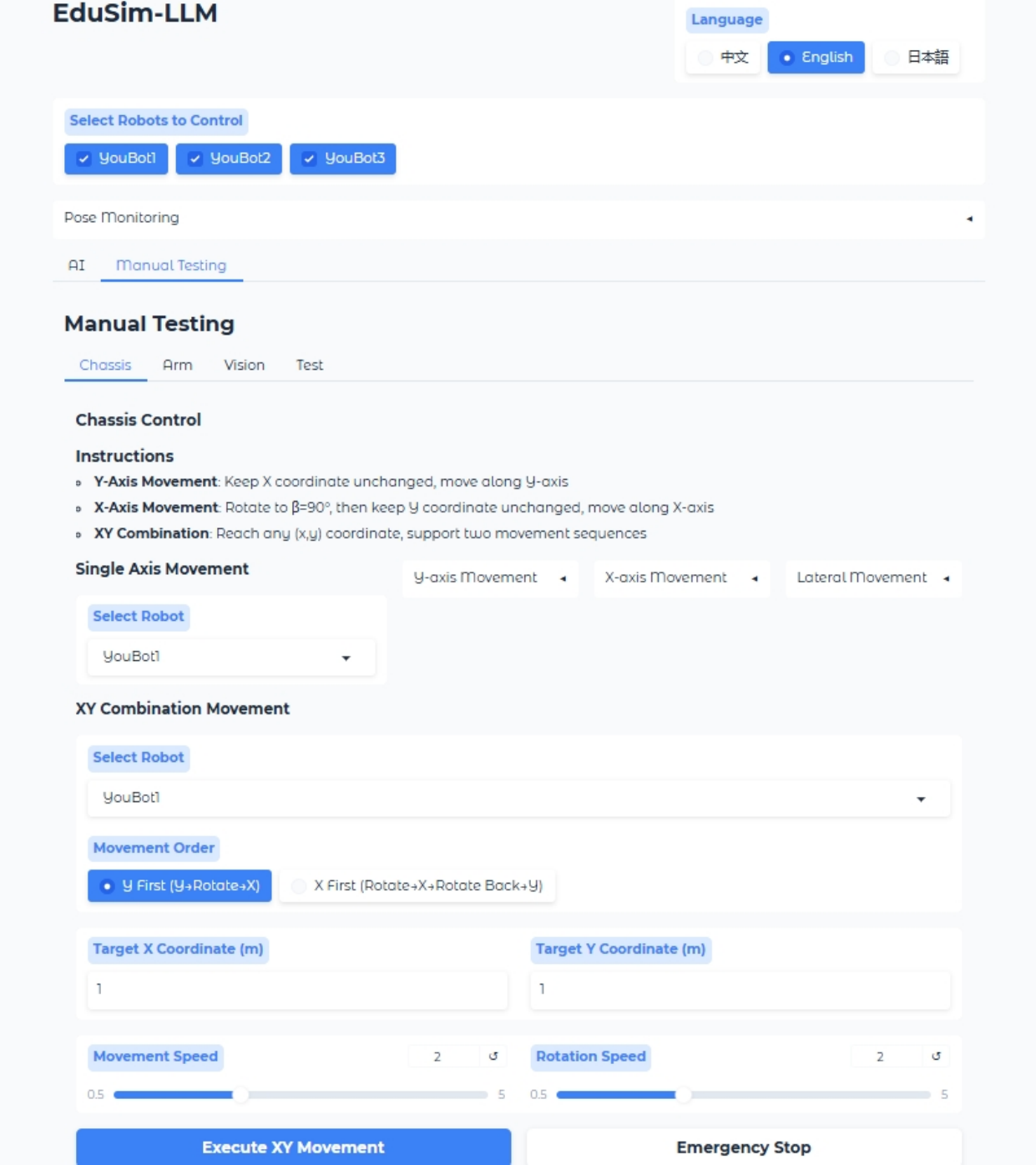}
  \caption{Mobile base manual motion testing interface}
  \label{fig: Frontend2}
\end{figure}

\begin{figure}[ht]
  \centering
  \includegraphics[width=1\textwidth]{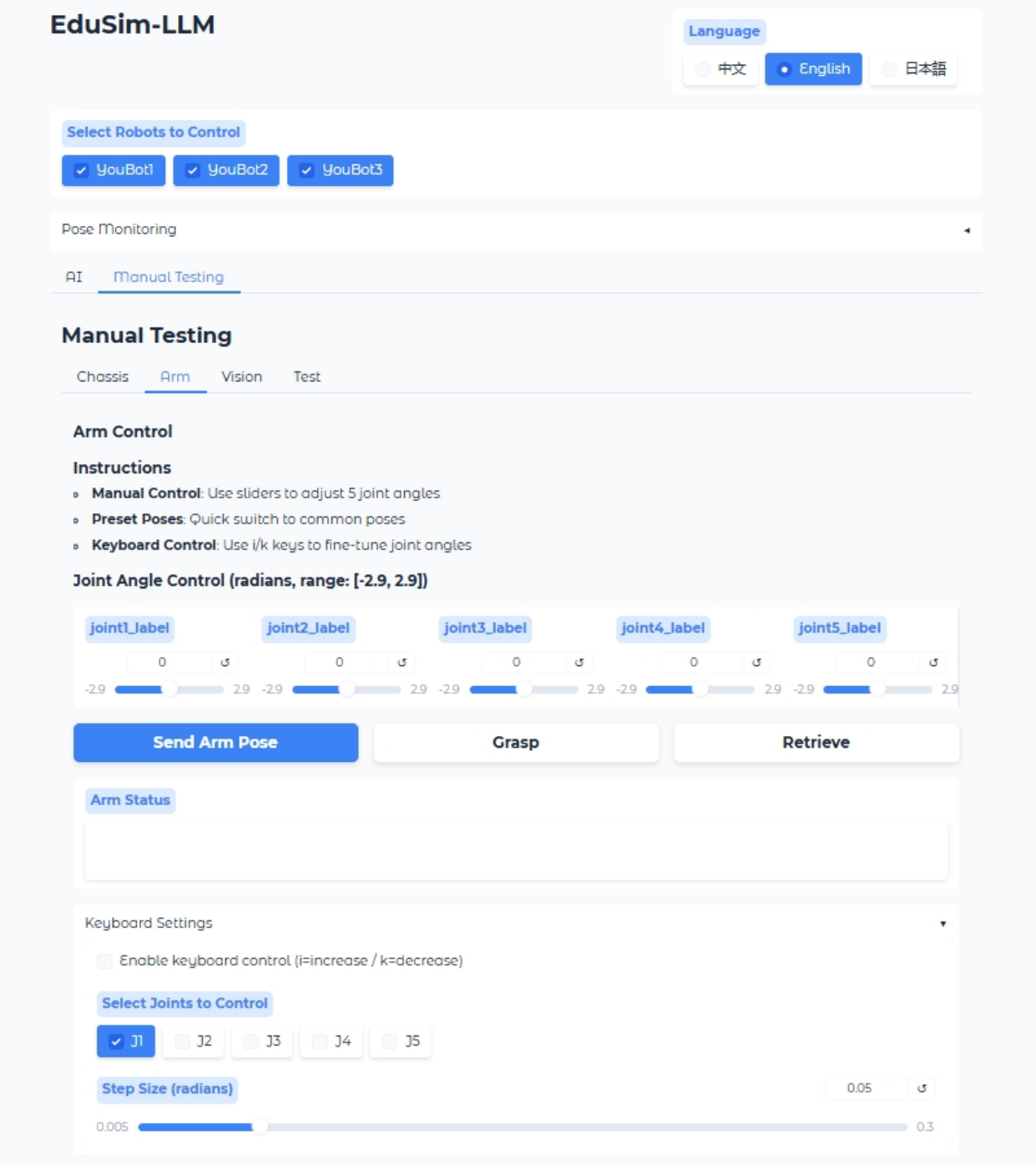}
  \caption{Manipulator and gripper manual control interface}
  \label{fig: Frontend3}
\end{figure}

\begin{figure}[ht]
  \centering
  \includegraphics[width=1\textwidth]{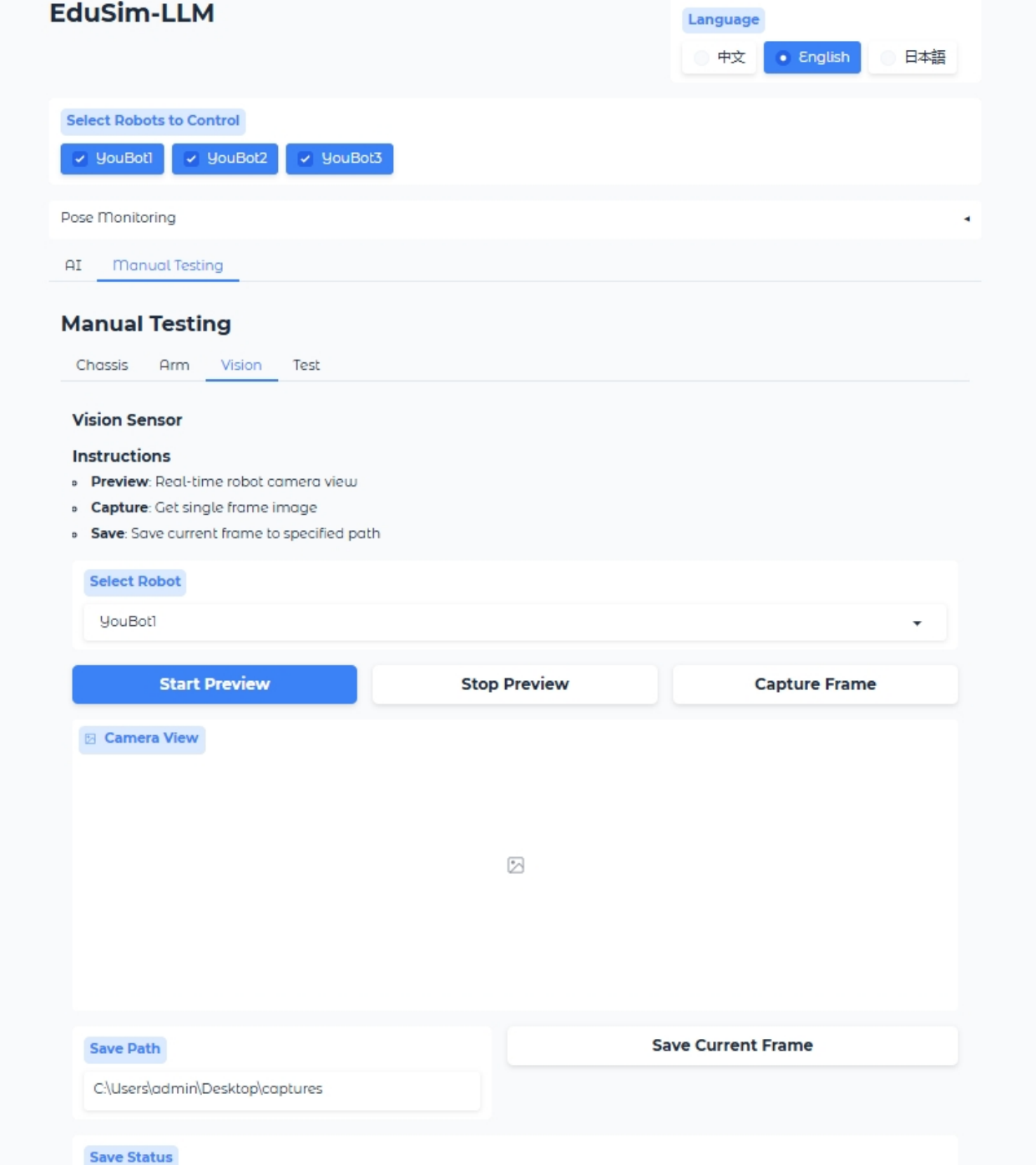}
  \caption{Vision sensor and image acquisition interface}
  \label{fig: Frontend4}
\end{figure}

Figs.~\ref{fig: Frontend1}--\ref{fig: Frontend4} show the EduSim-LLM frontend for command input and execution monitoring, and for manual control of the mobile base, manipulator/gripper, and vision-based image capture.

\end{document}